\documentclass[11pt,a4paper]{article}
\usepackage[hyperref]{naaclhlt2018}
\usepackage{times}
\usepackage{graphicx}
\usepackage{latexsym}
\pdfoutput=1

\usepackage{url}

\aclfinalcopy 

\title{Offensive Language Analysis using Deep Learning Architecture}

\author{Ryan Ong \\
  Faculty of Engineering, Department of Computing\\
  Imperial College London\\
  {\tt cmo18@ic.ac.uk}}

\date{}

\begin{document}
\maketitle
\begin{abstract}
SemEval-2019 Task 6 \cite{offenseval} requires us to identify and categorise offensive language in social media. In this paper we will describe the process we took to tackle this challenge. Our process is heavily inspired by \newcite{Sosa} where he proposed CNN-LSTM and LSTM-CNN models to conduct twitter sentiment analysis. We decided to follow his approach as well as further his work by testing out different variations of RNN models with CNN. Specifically, we have divided the challenge into two parts: data processing and sampling and choosing the optimal deep learning architecture. In preprocessing, we experimented with two techniques, SMOTE and Class Weights to counter the imbalance between classes. Once we are happy with the quality of our input data, we proceed to choosing the optimal deep learning architecture for this task. Given the quality and quantity of data we have been given, we found that the addition of CNN layer provides very little to no additional improvement to our model's performance and sometimes even worsen our F1-score. In the end, the deep learning architecture that gives us the highest macro F1-score is a simple BiLSTM-CNN.
\end{abstract}

\section{Introduction}
In this paper we will describe the process we took to tackle SemEval-2019 Task 6 \cite{offenseval}. \newcite{OLID} describes the dataset for this task. We have divided the challenge into two parts: data processing and sampling and choosing the optimal deep learning architecture. Given that our datasets are unstructured and informal text data from social medias, we have decided to spend more time creating our text preprocessing pipeline to ensure that we are feeding in high quality data to our model. In addition, we realised that there's a high level of imbalance between classes in each of the subtasks. Therefore, we decided to experiment with two different techniques that tackle this imbalance; SMOTE and Class Weights. 
Once our data is clean and our data distribution among classes are balanced, we proceed to choosing our optimal deep learning architecture. We decided to use macro F1-score as our evaluation metrics due to the imbalance classes. Through searching for the optimal model architecture, we made two important findings. Firstly, the order of our layering in our models heavily affects our F1-score performance. We found that by feeding data into the LSTM layer first, then followed by CNN layer yields much better results than the alternative. Secondly, in this challenge, the addition of CNN layer provides very little to no additional improvement and sometimes even lead to a decrease in our F1-score. We suspect that by feeding inputs into the CNN layer, we lose the important sequential information in text data, thereby making our models less accurate. In the end, we found that the deep learning architecture that gives us the highest F1-score among subtasks is BiLSTM-CNN.

\section{Deep learning architecture}
\begin{figure*}[tp]
\centering
\includegraphics[scale=0.5]{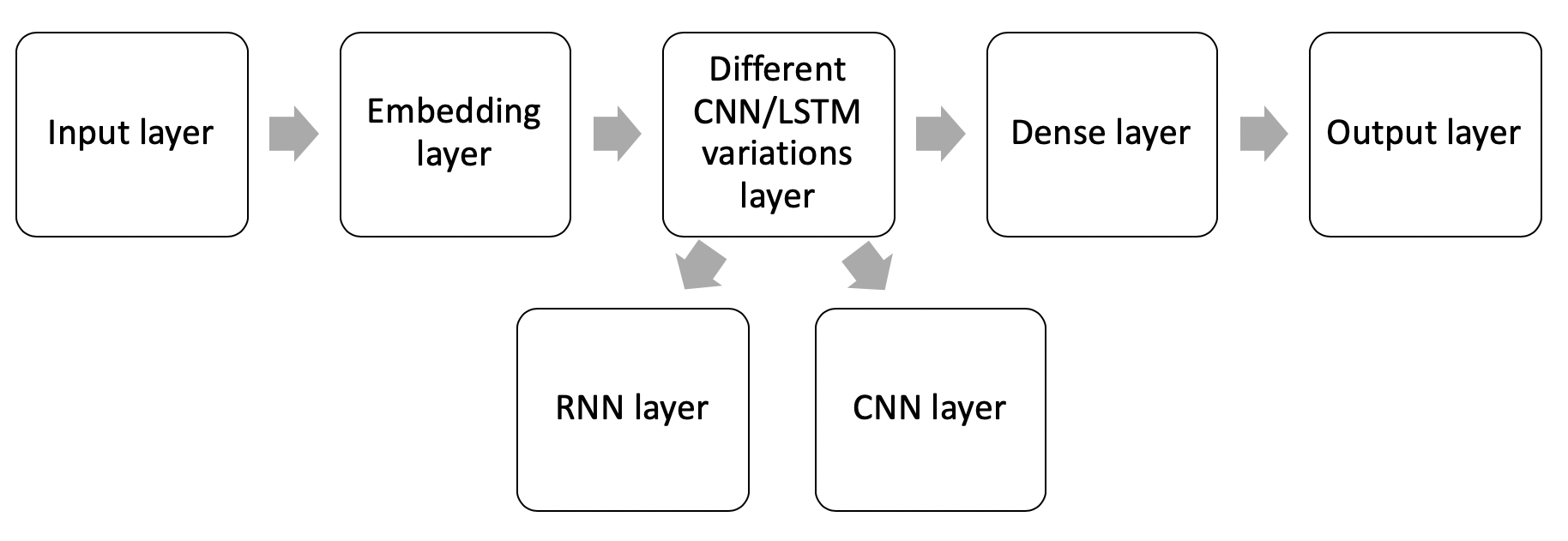}
\caption{Overall model architecture}
\label{Model Diagram}
\end{figure*}

In this paper, we experimented with different variations of CNN and LSTM layers. Our overall deep learning architecture is shown in Figure 1, where we initially feed our input text through an embedding layer to get our word embeddings. Depending on the variations of our CNN and LSTM layers, for example CNN-LSTM, we will feed these word embeddings to the convolution layer. The output will undergo MaxPooling layer (part of CNN), resulting in a smaller dimension output, which is then feed into the LSTM layer. We will then apply spatial dropout to the output of LSTM layer in an attempt to counter overfitting. This is followed by a dense layer before our model architecture outputs the results through the output layer.

The model was implemented using the Keras libary with Tensorflow backend.

\subsection{Pre-trained word embeddings}
Word embeddings are widely used in different NLP tasks. In this paper, we decided to experiment with different kinds of word embeddings, including different dimensionality of the same embeddings, to see if different type and dimensionality of embeddings would affect the overall end performance of our models. Specifically, we decided to experiment with the following word embeddings \cite{GloVe}:

\begin{enumerate}
	\item \textbf{\textit{GloVe: Twitter (100d)}} - Trained on 2B tweets, which is very relevant to our tasks of analysing text data from social media. It has 27B tokens and 1.2M vocabulary of unique words
	\item \textbf{\textit{GloVe: Twitter (200d)}} - Same as Glove: Twitter (100d) except it's 200 dimension. This allows us to evaluate the increase in dimensionality on the performance of our models
	\item \textbf{\textit{GloVe: Common Crawl (300d)}} - Trained on 42B tokens, 1.9M vocabulary of unique words. We chose this embeddings for its relatively large vocabulary size and embeddings dimensions
\end{enumerate}

\subsection{Optimisation and Regularisation}
Given our relatively small dataset, the network is trained using batch gradient descent with Adam optimiser. To counter overfitting, we have decided to utilise spatial dropout 1D regularisation which performs like a normal dropout regularisation except you drop the entire 1D feature maps instead of individual activation. This is because if adjacent frames within the same feature maps are highly correlated, then regular dropout will fail to regularise the activations.

\section{Training}
\subsection{Data}
To train and evaluate our models, we will be using the provided training and trial dataset. However, given the extremely small trial dataset, we have decided to combine both datasets as we aren't able to properly assess our models' predictions accurately with the trial dataset. Table 1 shows the label distribution of all the datasets. In addition, given the level of inbalance between classes in each subtasks, we have decided to focus more on the F1-scores, particularly the macro F1 score rather than just relying on the overall accuracy.

\begin{table*}[t!]
\centering
\begin{tabular}{|l|c|c|c|c|c|c|c|}
\hline
\multicolumn{1}{|c|}{} & \multicolumn{2}{c|}{Subtask A} & \multicolumn{2}{c|}{Subtask B} & \multicolumn{3}{c|}{Subtask C}\\
\cline{2-8}
\multicolumn{1}{|c|}{Dataset} & NOT & OFF & TIN & UNT & IND & GRP & OTH\\
\hline
\multicolumn{1}{|c|}{Train} & 8840 & 4400 & 3876 & 524 & 2407 & 1074 & 395\\
\multicolumn{1}{|c|}{Trial} & 243 & 77 & 38 & 39 & 30 & 4 & 5\\
\hline
Combined & 9083 & 4477 & 3914 & 563 & 2437 & 1078 & 400\\
\hline
\end{tabular}
\caption{Benchmark dataset label distribution}
\end{table*}

To train our model, we split the combined dataset randomly into 80\% train-val and 20\% test set and use the train-val set to perform k-fold cross validation (k = 5). Specifically we train each models using k-fold cross validation and use the validation set to do early stopping if the performance does not improve after 10 epochs with respect to average \textbf{macro F1-score}. Once we are happy with the performance of our final model, we do a final evaluation using the 20\% test set. 

\subsection{Preprocessing}
Our data preprocessing pipeline is as follows:
\begin{itemize}
\item \textit{Remove @USER and URL token}
\item \textit{Remove hashtags, twitter handles and hyperlinks}
\item \textit{Apostrophe contraction-to-expansion} - We used a dictionary to map contracted words to their corresponding expanded words. For example, \textit{don't} will transform to \textit{do not}. This preprocessing steps reveal "hidden" negation words that are important for our models to detect offensive languages
\item \textit{Spelling corrections} - We used open-source Sympell \cite{SymSpell} which uses the Damerau-Levenshtein distance to find the closest correct spellings for any misspelled words. We chose the edit distance to be 3
\item \textit{Lemmatisation} - We used WordNetLemmatizer from NLTK to lemmatise all the words to their lemma form. For example, \textit{saw} to \textit{see}. By lemmatising the words, we only feed in words in their lemma form, therefore allowing the models to be able to capture the meaning of words regardless of their original forms.
\item \textit{All text are lowercased}
\end{itemize}

\subsection{Class Imbalance}
The provided dataset has a high level of class imbalance (shown in Table 1) and we have decided to use two different approaches to counter this: \textbf{class weights} and \textbf{SMOTE} \cite{smote_paper}. Class weights involves computing the class weights and use it to re-scale the loss function when performing back-propagation. SMOTE, on the other hand, is an oversampling technique whereby it generates new data points using the existing minority data that we supply as input. The algorithm takes samples of the feature space for each target class and its nearest neighbors and generates new examples that combine features of the target case with features of its neighbors \cite{SMOTE_def}.

\section{Experiments and results}
\subsection{Experiment environment}
In order to find the optimal architecture for this task, we have decided to experiment with CNN and different variations of RNN, which includes LSTM and GRU (each either unidirectional or Bidirectional). Each variation of model will follow the same overall structure as mentioned in Section 2.

We will be making performance comparisons between the models below:

\begin{enumerate}
	\item CNN
	\item LSTM
	\item BiLSTM
	\item GRU
	\item BiGRU
	\item CNN-LSTM
	\item CNN-BiLSTM
	\item CNN-GRU
	\item CNN-BiGRU
	\item LSTM-CNN
	\item BiLSTM-CNN
	\item GRU-CNN
	\item BiGRU-CNN
\end{enumerate}

Each model will be train using 5-fold cross validation and will be evaluated on the average accuracy and macro F1-score. 

\subsection{Results analysis - Subtask A}
The results on Table 2 shows the average accuracy and macro F1-score of each architecture after 5-fold cross validation.

\begin{table*}[t!]
\centering
\begin{tabular}{|r|c|c|}
\hline Models (Subtask A) & Avg Acc & Avg Macro F1 \\ \hline
CNN & 71\% & 0.63 \\
\textbf{LSTM} & \textbf{78\%} & \textbf{0.73} \\
\textbf{BiLSTM} & \textbf{78\%} & \textbf{0.73} \\
\textbf{GRU} & \textbf{79\%} & \textbf{0.74} \\
\textbf{BiGRU} & \textbf{78\%} & \textbf{0.74} \\
CNN-LSTM & 71\% & 0.66\\
CNN-BiLSTM & 72\% & 0.65 \\
CNN-GRU & 72\% & 0.66 \\
CNN-BiGRU & 72\% & 0.67 \\
LSTM-CNN & 74\% & 0.67 \\
\textbf{BiLSTM-CNN} & \textbf{78\%} & \textbf{0.74} \\
GRU-CNN & 73\% & 0.67 \\
\textbf{BiGRU-CNN} & \textbf{77\%} & \textbf{0.72} \\
\hline
\end{tabular}
\caption{\label{font-table} Average accuracy and macro F1-score of different model architecture (k-fold = 5) - \textbf{Subtask A}}
\end{table*}

Given our inbalanced datasets, we will primarily be evaluating our models using the average macro F1-score. A standalone CNN model yields the lowest average macro F1-score of 0.63. Through adding an LSTM or GRU (either unidirectional or bidirectional) layer after the CNN layer, thereby forming a LSTM-CNN or GRU-CNN, our model scores on average 0.02 - 0.04 higher than standalone CNN model. However, this is 0.06 lower than standalone LSTM (0.73). A possible reason for this could be that although CNN layer is great at extracting local features and learn to emphasise or disregard certain n-grams in the input data, it still looses some of the important sequential information in our text input.

On the other hand, a standalone LSTM or GRU model yields the highest average macro F1-score of 0.73-0.74. Our results show that there's no significant difference between unidirectional and bidirectional LSTM or GRU. Intuitively, the benefit of a LSTM or GRU layer is that the network will be able to remember what was read previously, therefore can develop a better understanding of future inputs. We found that a normal unidirectional LSTM-CNN or GRU-CNN underperformed relatively to standalone LSTM/GRU models and only outperforms standalone CNN marginally by 0.04. BiLSTM-CNN/BiGRU-CNN achieve average macro F1-score similar to standalone LSTM/GRU. Our results show that adding a CNN layer after LSTM/GRU provides no benefits or worsen the score.

Overall, our results show that the ordering of layers significantly affect the performance of our models. Our results indicate that the optimal ordering of layers is LSTM/GRU follow by CNN, thereby forming a LSTM-CNN/GRU-CNN architecture. The initial LSTM/GRU layer will be able to capture sequential information unlike having CNN layer as the first layer. The output is then pass to the CNN layer to extract local features.

\subsection{Subtask B and C}

\begin{table*}[t!]
\centering
\begin{tabular}{|l|c|c|c|c|c|c|}
\hline
\multicolumn{1}{|c|}{Models} & \multicolumn{2}{c|}{Imbalanced Data} & \multicolumn{2}{c|}{SMOTE} & \multicolumn{2}{c|}{Class Weights}\\
\cline{2-7}
\multicolumn{1}{|c|}{(Subtask B)} & Acc & Macro F1 & Acc & Macro F1 & Acc & Macro F1\\
\hline
\multicolumn{1}{|c|}{BiLSTM-CNN} & 87.39\% & 0.51 & 81.25\% & \textbf{0.59} & 68.86\% & 0.55\\
\multicolumn{1}{|c|}{BiGRU-CNN} & 87.83\% & 0.47 & 78.68\% & \textbf{0.57} & 44.64\% & 0.40\\
\multicolumn{1}{|c|}{BiLSTM} & 87.50\% & \textbf{0.56} & 79.80\% & 0.53 & 43.86\% & 0.40\\
\multicolumn{1}{|c|}{BiGRU} & 87.95\% & 0.53 & 75.22\% & \textbf{0.55} & 55.24\% & 0.48\\
\hline
\end{tabular}
\caption{Evaluation of different techniques to tackle class imbalance. Table displays accuracy and macro F1-score of different model architecture (holdout method) - \textbf{Subtask B}}
\end{table*}

\begin{table*}[t!]
\centering
\begin{tabular}{|l|c|c|c|c|c|c|}
\hline
\multicolumn{1}{|c|}{Models} & \multicolumn{2}{c|}{Imbalanced Data} & \multicolumn{2}{c|}{SMOTE} & \multicolumn{2}{c|}{Class Weights}\\
\cline{2-7}
\multicolumn{1}{|c|}{(Subtask C)} & Acc & Macro F1 & Acc & Macro F1 & Acc & Macro F1\\
\hline
\multicolumn{1}{|c|}{BiLSTM-CNN} & 69.99\% & \textbf{0.48} & 66.16\% & 0.45 & 59.13\% & 0.44\\
\multicolumn{1}{|c|}{BiGRU-CNN} & 71.14\% & 0.42 & 68.20\% & \textbf{0.45} & 63.09\% & 0.35\\
\multicolumn{1}{|c|}{BiLSTM} & 69.48\% & \textbf{0.45} & 67.82\% & 0.45 & 61.30\% & 0.45\\
\multicolumn{1}{|c|}{BiGRU} & 71.39\% & \textbf{0.46} & 64.11\% & 0.43 & 62.58\% & 0.43\\
\hline
\end{tabular}
\caption{Evaluation of different techniques to tackle class imbalance. Table displays accuracy and macro F1-score of different model architecture (holdout method) - \textbf{Subtask C}}
\end{table*}

Given our findings on the optimal ordering of layers and the fact that BiLSTM-CNN and BiGRU-CNN significantly outperformed normal LSTM-CNN and GRU-CNN in subtask A, we have decided to only apply BiLSTM, BiGRU, BiLSTM-CNN and BiGRU-CNN to subtask B and C. The holdout results for subtask B and C are shown in Table 3 and 4 respectively. We decided not to use cross validation for subtask B and C due to computationally intensive to run. The results show that SMOTE is the best technique to tackle the class imbalance issue. With the exception of BiLSTM, we performed top macro F1-score for the other three models. 
However, for subtask C, it seems that our results has got worse since applying SMOTE/Class Weights to the datasets, with the exception of BiGRU-CNN. Our results indicate that it is better off keeping the original datasets.

Taking the results from our experiments, we conclude that the optimal deep learning architecture to tackle SemEval-2019 Task 6 offensive language analysis is BiLSTM-CNN as it consistently outperforms every other model variations. We decided to not apply SMOTE or class weights to datasets in subtask A as the level of imbalance in subtask A is mild. In terms of subtask B and C, it is clear that we should apply SMOTE to balance our data among classes in order to yield the highest possible macro F1-score.

\subsection{Hyperparamter Tuning and Findings}
Once we finalise our model to be BiLSTM-CNN, we conducted manual search for some of the key hyperparameters of the model using subtask A. This include optimal number of epochs to train our model, the spatial dropout probability and the use of different types and dimensions of word embeddings. We included BiGRU-CNN as a comparison. The results is as follows:
\begin{enumerate}
	\item \textbf{Level of Epochs} - As shown in Table 5, the optimal number of epochs to train our model is 5. Our model managed to reach high F1-score of 0.74/0.75 (relative to our experiments) after 5 epochs. The F1-score starts to plateau/drop as we increase the number of epochs beyond 5, showing signs of overfitting
	
    \begin{table}[tp]
    \centering
    \begin{tabular}{|c|c|c|}
    \hline Epochs & BiLSTM-CNN & BiGRU-CNN \\ \hline
    5 & \textbf{0.74} & \textbf{0.75}\\
    10 & 0.70 & 0.70\\
    20 & 0.71 & 0.73\\
    \hline
    \end{tabular}
    \caption{\label{font-table} Macro F1-score for BiLSTM-CNN trained with different epochs - \textbf{Subtask A}}
    \end{table}

	\item \textbf{Spatial dropout rate} - We have spatial dropout layer immediately after the output of our BiLSTM-CNN model as well as after the dense layer (Figure 1). As shown in Table 6, the optimal spatial dropout rate is 20\%. However, when taken out the spatial dropout layer, our macro F1-score was not affected. This might be due to our small network architecture and low overfitting, therefore dropout layer doesn't contribute much to our final performance
	
    \begin{table}[tp]
    \centering
    \begin{tabular}{|c|c|c|}
    \hline Dropout & BiLSTM-CNN & BiGRU-CNN \\ \hline
    20\% & \textbf{0.75} & 0.73\\
    35\% & 0.74 & 0.70\\
    50\% & 0.72 & 0.72 \\
    No Dropout & 0.74 & \textbf{0.74}\\
    \hline
    \end{tabular}
    \caption{\label{font-table} Macro F1-score for BiLSTM-CNN trained with different spatial dropout rates - \textbf{Subtask A}}
    \end{table}
    
	\item \textbf{Pre-trained vs No pre-trained embeddings} - Our results in Table 7 aligns with the industry trend that by using pre-trained word embeddings, we yield a higher macro F1-score when compared trained without pre-trained word embeddings. In addition, we see an increase in the performance of our BiLSTM-CNN as we increased the dimensions of our word embeddings. However, due to the contrasting results from BiGRU-CNN, we aren't unable to draw a conclusion and further experiments is needed
	
    \begin{table}[tp]
    \centering
    \begin{tabular}{|p{1.8cm}|p{2.34cm}|p{2.12cm}|}
    \hline Embeddings & BiLSTM-CNN & BiGRU-CNN \\ \hline
    \multicolumn{1}{|c|}{T - 100d} & \multicolumn{1}{|c|}{0.72} & \multicolumn{1}{|c|}{\textbf{0.74}}\\
    \multicolumn{1}{|c|}{T - 200d} & \multicolumn{1}{|c|}{\textbf{0.75}} & \multicolumn{1}{|c|}{0.73}\\
    \multicolumn{1}{|c|}{CC - 300d} & \multicolumn{1}{|c|}{\textbf{0.75}} & \multicolumn{1}{|c|}{0.71}\\
    \multicolumn{1}{|c|}{No Embs} & \multicolumn{1}{|c|}{0.69} & \multicolumn{1}{|c|}{0.67}\\
    \hline
    \end{tabular}
    \caption{\label{font-table} T - GloVe Twitter, CC - GloVe Common Crawl. Macro F1-score for BiLSTM-CNN trained with/without pre-trained embeddings - \textbf{Subtask A}}
    \end{table}

\end{enumerate}

\section{Conclusion}

From all our experiments, we concluded that our optimal model architecture is BiLSTM-CNN, trained with 5 epochs, no dropout layers (unless we decided to build a bigger model architecture) and use of pre-trained word embeddings, 42B GloVe-Common Crawl (300d). In this paper, we experimented with 13 model variations with the aim to find the optimal model architecture for offensive language analysis. Our findings show that the ordering of layers in our model are extremely important. By having CNN layer first followed by different types of RNN layers, our models perform 0.07 - 0.09 worse in terms of F1-score when compared to having RNN layers first followed by a CNN layer. We used BiLSTM-CNN to predict the labels for the hidden test set and our final macro F1-scores and rankings are shown in Table 8. \textbf{Our code is available at:} https://github.com/RyanOngAI/semeval-2019-task6

\begin{table}
\centering
\begin{tabular}{|c|c|c|}
\hline Subtasks & Macro F1 & Ranking \\ \hline
A & \textbf{0.75} & 56\\
B & \textbf{0.65} & 38\\
C & \textbf{0.46} & 77\\
\hline
\end{tabular}
\caption{\label{font-table} Macro F1-score \& Ranking - Hidden test set}
\end{table}

\subsection{Future Work}
\begin{enumerate}
	\item \textbf{Systematic search} - Manual hyperparameter search limits the number of experiments I can carry out, for example, I wasn't able to manually test out different dropout and recurrent dropout rate within the RNN layers. This has been set to 35\% randomly. Therefore it would be beneficial to implement different systematic search such as grid search or bayesian optimisation to optimise the hyperparameters for our models
	\item \textbf{Contextualised word embeddings} - On top of tradition word embeddings, it would also be interesting to see how contextualised embeddings would affect the results of our models given the rise of BERT and ELMO
	\item \textbf{Character-level} - Given the informal nature of our text data, it would be interesting to see the results of character level model variations of our experiments above seeing as the full power of pre-trained word embeddings is limited by the misspelled/slang words
\end{enumerate}

\end{document}